\documentclass[conference]{IEEEtran}
\IEEEoverridecommandlockouts
\usepackage{cite}
\usepackage{amsmath,amssymb,amsfonts}
\usepackage{graphicx}
\usepackage{textcomp}
\usepackage{xcolor}
\def\BibTeX{{\rm B\kern-.05em{\sc i\kern-.025em b}\kern-.08em
    T\kern-.1667em\lower.7ex\hbox{E}\kern-.125emX}}
    
\usepackage{algorithm}
\usepackage{algorithmicx}
\usepackage{algpseudocode}
    
\usepackage{adjustbox}
\usepackage{caption}
\usepackage{microtype}
\usepackage{booktabs}
\usepackage{multirow} 
\usepackage{marvosym}
\usepackage{cite}
\usepackage{multirow}

\usepackage{amsthm}

\usepackage{listings}
\usepackage{ifxetex}
\usepackage{fancyhdr}
\usepackage{hyperref}
\usepackage{graphicx}
\usepackage{wrapfig}
\usepackage{amsmath}
\usepackage{amsfonts}
\usepackage{amssymb}
\usepackage{listings}
\usepackage{setspace}
\usepackage{wrapfig}
\usepackage{tablefootnote}
\usepackage{multirow}
\usepackage{booktabs}
\usepackage[section]{placeins}
\usepackage{color}
\usepackage{amsfonts}
\usepackage{ulem}
\usepackage{float}
\usepackage{dsfont}
\usepackage{cleveref}

\newcommand{\set}[1]{\mathcal{#1}}
\newcommand{\pnorm}[1]{\lVert{#1}\rVert}

\DeclareMathOperator*{\loss}{{\ell}}

\newcommand{\RN}{\mathbb{R}}

\newcommand{\refeq}[1]{Eq.~\eqref{#1}}

\DeclareMathOperator*{\diversity}{\ensuremath{{\psi}}}
\newcommand{\x}{\ensuremath{\vec{x}}}

\newcommand{\y}{\ensuremath{y}}

\newcommand{\setX}{\ensuremath{\set{X}}}
\newcommand{\setY}{\ensuremath{\set{Y}}}

\newcommand{\setF}{\ensuremath{\set{F}}} 
\newcommand{\xcf}{\ensuremath{\vec{x}_{\text{cf}}}}
\newcommand{\xsf}{\ensuremath{\vec{x}_{\text{sf}}}}

\newcommand{\deltasf}{\ensuremath{\vec{\delta}}}

\newcommand{\dimsym}{d}

\newcommand{\classifier}{\ensuremath{h}}

\newcommand{\reject}{\ensuremath{r}}
\newcommand{\threshold}{\ensuremath{\theta}}

\newcommand{\mySF}[3]{\ensuremath{\text{SF}_{#3}(#1, #2)}}

\newcommand{\rejectSymbol}{\infty}
\newcommand{\xnew}{\ensuremath{\x_{*}}}

\newcommand{\Dcalib}{\ensuremath{\set{D}_\text{calib}}}

\newcommand{\nonconformity}{\ensuremath{\phi}}
\newcommand{\credibility}{\ensuremath{\psi}}

\begin{document}

\title{``Even if ...'' -- Diverse Semifactual Explanations of Reject
\thanks{We gratefully acknowledge funding from the VW-Foundation for the project \textit{IMPACT} funded in the frame of the funding line \textit{AI and its Implications for Future Society}.

Andr\'e Artelt is also affiliated with KIOS Research and Innovation Center of Excellence, University of Cyprus, Nicosia, Cyprus.
}
}

\author{\IEEEauthorblockN{1\textsuperscript{st} Andr\'e Artelt}
\IEEEauthorblockA{\textit{CITEC - Faculty of Technology} \\
\textit{Bielefeld University}\\
Bielefeld, Germany \\
aartelt@techfak.uni-bielefeld.de}
\and
\IEEEauthorblockN{2\textsuperscript{nd} Barbara Hammer}
\IEEEauthorblockA{\textit{CITEC - Faculty of Technology} \\
\textit{Bielefeld University}\\
Bielefeld, Germany \\
bhammer@techfak.uni-bielefeld.de}
}

\maketitle

\begin{abstract}
Machine learning based decision making systems applied in safety critical areas require reliable high certainty predictions. For this purpose, the system can be extended by an reject option which allows the system to reject inputs where only a prediction with an unacceptably low certainty would be possible. While being able to reject uncertain samples is important, it is also of importance to be able to explain why a particular sample was rejected. With the ongoing rise of eXplainable AI (XAI), a lot of explanation methodologies for machine learning based systems have been developed -- explaining reject options, however, is still a novel field where only very little prior work exists.

In this work, we propose to explain rejects by \textit{semifactual explanations}, an instance of example-based explanation methods, which them self have not been widely considered in the XAI community yet. We propose a conceptual modeling of semifactual explanations for arbitrary reject options and empirically evaluate a specific implementation on a conformal prediction based reject option.
\end{abstract}


\section{Introduction}
The usage of machine learning (ML) based systems in safety critical areas such as autonomous driving~\cite{AutonmousDriving} requires high certainty predictions -- i.e. making mistakes must be avoided at all costs because those might have serious consequences. However, most ML models do not provide proper and reliable certainty scores of their predictions, and the models usually always output a prediction no matter how certain this prediction and how reasonable the input is. A potential remedy is to extend ML models with \textit{reject options}~\cite{hendrickx2021machine}. Reject options allow the model to reject inputs (i.e. refuse to make a prediction) where the model is not certain in its prediction -- by this it can be ensured that we only make high certainty predictions and avoid making serious mistakes. For instance consider the example of a phishing mail filter: \textit{Imagine a mail filtering application that tries to filter out phishing mails in order to protect the end users and their surrounding from serious consequences. The filter is supposed to automatically sort out mails where it it is certain that the particular mails are malicious, and pass all benign mails to the user without any delay. However, in cases where the filter is not absolute certain about its prediction (distinguishing benign vs. malicious), it should reject to classify this mail -- such rejected mails might be passed to the user with an additional warning of taking care or to the IT-security department of the company for further investigations and possible improvement of the filtering application.}

Because of the increasing usage of ML based systems in practice, transparency of such systems is nowadays a widely accepted standard requirement which also made its way into legal regulations like the EU's GDPR~\cite{GDPR}. Transparency is usually realized by means of explanations -- i.e. explanations of the system's behavior are provided to the user~\cite{molnar2019}. Although it is still not perfectly understood what exactly makes up  ``good'' explanations~\cite{doshivelez2017rigorous,offert2017i}, there have been developed a wide variety of different explanation methods~\cite{ExplainingBlackboxModelsSurvey,molnar2019,SurveyXai}: Feature relevance/importance methods~\cite{FeatureImportance} and examples based methods~\cite{CaseBasedReasoning}. Instances of example based methods are contrasting explanations like counterfactual explanations~\cite{counterfactualwachter,CounterfactualReviewChallenges} and prototypes \& criticisms~\cite{PrototypesCriticism} -- these methods use a set or a single example for explaining the behavior of the system.

Besides being able to reject inputs that lead to low-certainty predictions, it is also important to be able to explain why some inputs got rejected while others got accepted -- in order to trust the reject option, the user has to get an understanding on how this particular reject option works. E.g. coming back to the previous example: \textit{If the filtering application rejects to classify a mail as either benign or malicious, it would be beneficial to get an explanation why the filtering application was not able to come up with a high certainty prediction -- in particular when it comes to understand the reasons for this behavior and potential improvements of the filtering application.}

\paragraph*{Related Work}
Despite the obvious importance of explaining reject options in ML systems, surprisingly little work exists on this topic:

The authors of~\cite{artelt2022explaining} propose to use counterfactual explanations for explaining reject options of learning vector quantization models -- i.e. there approach is limited to this particular type of model and reject option. Furthermore, only a single explanation is computed -- note that according to the Rashomon effect there might exist more than one possible explanation, and therefore a lot of information might be lost by just computing and presenting a single explanation.

Another work~\cite{artelt2022model} proposes a model agnostic approach for locally explaining arbitrary reject options of arbitrary systems -- they propose to use a local and interpretable approximation (e.g. a linear model or decision tree) to locally explain the reject option. However, a local approximation might not always capture the entire behavior of the global function and it is also debatable whether a linear model or a decision tree is always truly ``interpretable''.

\paragraph*{Our Contributions}
Our contributions in this work is of conceptional nature: We propose a model agnostic method for explaining reject options by means of a set of diverse semifactual explanations -- we propose the abstract concept, a formalization of this concept, and an implementation of it.
We not only consider an explanation method that has not been considered before for explaining reject options, we also compute a set of diverse explanation to maximize the amount of information that is presented to the user -- i.e. dealing with the ``Rashomon effect'' which states that there often exist more than one possible explanation.

The remainder of this work is structured as follows: First, we introduce the necessary foundations of this work: we review reject options in Section~\ref{sec:foundations:rejectoptions} and semifactual explanations in Section~\ref{sec:foundations:semifactuals}. Next in Section~\ref{sec:modeling}, we introduce our model agnostic modeling for computing diverse semifactual explanations of reject options. We empirically evaluate quantitative aspects of our proposed diverse semifactual explanations of reject options on several standard benchmark data sets in Section~\ref{sec:experiments}. Finally, this work closes with a conclusion in Section~\ref{sec:conclusion}.

\section{Foundations}\label{sec:foundations}
\subsection{Reject Options}\label{sec:foundations:rejectoptions}
For a given prediction function $\classifier:\setX\to\setY$, a reject option~\cite{hendrickx2021machine} is usually added by means of a certainty function $\reject_{\classifier}:\setX\to\RN_{+}$ that measures the certainty of the prediction $\classifier(\x)$ -- we reject a sample $\x$ if the certainty is below a given threshold $\threshold$:
\begin{equation}\label{eq:rejectoption}
	\reject_{\classifier}(\x) < \threshold
\end{equation}
where the subscript $\classifier$ denotes a potential dependency on the prediction function $\classifier(\cdot)$.

We can think about adding a reject option to $\classifier(\cdot)$ as constructing a new function $\classifier':\setX\to\setY\cup\{\rejectSymbol\}$ where we add a reject symbol $\rejectSymbol$ to the set of possible outputs $\setY$:
\begin{equation}\label{eq:classifier_with_rejectoption}
    \classifier'(\x) =
    \begin{cases}
    \classifier(\x)       & \quad \text{if } \reject_{\classifier}(\x) \geq \threshold\\
    \rejectSymbol  & \quad \text{otherwise}
  \end{cases}
\end{equation}
There exist a lot of different implementation of certainty based reject options~\refeq{eq:rejectoption}~\cite{hendrickx2021machine}. In this work, we use the same conformal prediction based reject option~\cite{linusson2018classification} as it was done in~\cite{artelt2022model}:
Assume a (black-box) probabilistic classifier $\classifier:\setX\to\setY$ $\classifier(\x) = \underset{\y\,\in\,\setY}{\arg\max}\;p(\y \mid \x)$,
where $p(\y \mid \x)$ denotes the class wise probability as estimated by the classifier $\classifier(\cdot)$.
The so called non-conformity measure $\nonconformity_{\classifier}:\setX,\setY\to\RN$ -- which measures how different a given labeled sample is from a given set of labeled samples we have seen before -- is a central building block of a conformal predictor~\cite{ConformalShaferV08}.
In case of a probabilistic classifier $\classifier(\cdot)$, a common non-conformity measure is given as follows:
$\nonconformity_{\classifier}(\x,\y=j) = \underset{i\neq j}{\max}\; p_{\classifier}(\y=i\mid\x) - p_{\classifier}(\y=j\mid\x)$.
For fitting a conformal predictor based on $\classifier(\cdot)$, we need another labeled data set $\Dcalib\subset\setX\times\setY$ (called calibration set) which was not used during the fitting of $\classifier(\cdot)$ -- 
we compute the non-conformity $\alpha_i$ of every sample from the calibration set by applying $\nonconformity_{\classifier}(\cdot)$:
$\alpha_i = \nonconformity_{\classifier}(\x_i, \y=\y_i)$.

For every new data point $\xnew\in\setX$ that has to be classified, we compute the non-conformity measure for every possible label in $\setY$.
Next, the non-conformity scores of $\xnew$ are compared with the non-conformity scores from the calibration set to compute p-values $p_{\y=i}(\cdot)$ for every possible classification of $\xnew$.
The conformal predictor then selects the label with the larges p-value as a prediction $\classifier(\xnew) = \underset{i\,\in\,\setY}{\arg\max}\;p_{\y=i}(\xnew)$.
The credibility of the prediction  -- i.e. how well the training set supports the prediction -- is given by the largest p-value: $\credibility(\xnew) = \underset{i}{\max}\;p_{\y=i}(\xnew)$.

We implement a reject option~\refeq{eq:rejectoption}~\cite{linusson2018classification} using the credibility score:
\begin{equation}\label{eq:reject:credibility}
    \reject_{\classifier}(\x) = \credibility(\x) = \underset{i}{\max}\;p_{\y=i}(\x)
\end{equation}

\subsection{Semifactual Explanations}\label{sec:foundations:semifactuals}
A semifactual explanation (often just called \textit{semifactual}) states some changes that even if they had been applied to the input, would \textit{not change} the models behavior/prediction. We can think of a semifactual explanation as an \textit{``Even if ...''} explanation.
Consider the popular example of local application~\cite{CreditRiskML,molnar2019}. \textit{Imagine that a loan application was rejected and we want to know why. A semifactual explanation could be that the application would have been still rejected, even if the applicant had earned 500\$ more per month and even if they had not have a second credit card}.

Note that in contrast to contrasting explanations like counterfactual explanations~\cite{counterfactualwachter}, a semifactual does not recommend actions how to change the system's behavior -- this is what a contrasting explanation would do -- but it rather tells the user what changes would not change the system's behavior. A semifactual can be interpreted in at least two different ways: A semifactual tries to convince the user that the system's behavior can not be changed (easily) -- this could be kind of an ``evil'' use because it creates the impression that there is no way that the system's behavior could be changed and therefore the user must simply accept it. Another interpretation could be that a semifactual tells the user where not to put effort into because it does not help to change the system's behavior -- for improved usefulness, this could be enriched with a contrasting explanation recommending some actions to the user for how to change the system's behavior. In this case, one could also interpret a semifactual explanation as kind of an explanation of the contrasting explanation -- i.e. why it must be done as suggested in the contrasting explanation and not somehow different.

From a computer science perspective, an interesting question is how to exactly model the intuition of a semifactual explanation and how to compute these kinds of explanation efficiently.
While semifactual explanations haven been studied for quite some time and are well known in philosophy~\cite{FactFictionForecast,EvenIf,EvenStillCounterfactuals} and psychology~\cite{CounterfactualThinking,CounterfactualSemifactual,CounterfactualSemifactualThinking}, there are -- to the best of our knowledge --  more or less unknown in the ML community. The only work, we are away of is~\cite{PlausibleSemifactualsDeep} in which the authors study how to compute realistic semifactual explanation of deep neural networks applied to computer vision problems (in particular image classification).

\section{Semifactual Explanations of Reject}\label{sec:modeling}
\begin{algorithm}[t]
\caption{Computation of Diverse Semifactuals}\label{algo:diverse_sf}
\textbf{Input:} Original input $\x$, $k\geq 1$: number of diverse semifactuals
, Reject option $\reject(\cdot)$ \\
\textbf{Output:} Set of diverse semifactuals $\set{R}=\{\xsf^i\}$
\begin{algorithmic}[1]
 \State $\setF = \{\}$  \Comment{Initialize set of black-listed features}
 \State $\set{R} = \{\}$  \Comment{Initialize set of diverse semifactuals}
 \For{$i=1,\dots,k$} \Comment{Compute $k$ diverse semifactuals}
    \State $\xsf^i = \mySF{\x}{\set{F}}{\reject(\cdot)}$  \Comment{Compute next semifactual}
    \State $\set{R} = \set{R} \cup \{\xsf^i\}$
    \State $\setF = \setF \cup \{j \mid (\xsf - \x)_j \neq 0\}$  \Comment{Update set of black-listed features}
 \EndFor
\end{algorithmic}
\end{algorithm}
In this Section we propose a model agnostic modeling of (diverse) semifactual explanations (see Section~\ref{sec:foundations:semifactuals}) of reject options -- i.e. our proposed modeling is not tailored towards a specific reject option or model.

We formalize a single semifactual explanation as an solution to an optimization problem:
\begin{equation}\label{eq:sf:opt}
    \underset{\xsf\,\in\RN^\dimsym}{\arg\min}\;\loss(\xsf)
\end{equation}
where the loss function $\loss(\xsf)$ consists of different parts covering different aspects of semifactual explanations:
\begin{equation}\label{eq:sf:loss}
    \loss(\xsf) = {\loss}_{\text{feas-sf}}(\xsf) + {\loss}_\text{diverse}(\xsf) + {\loss}_\text{similarity}(\xsf) + {\loss}_\text{simple}(\xsf)
\end{equation}
Since~\refeq{eq:sf:loss} might not be differentiable for arbitrary models, we might have to solve~\refeq{eq:sf:opt} by using a general black-box methods such as Downhill-Simplex method.

In the following, we introduce and motivate the different parts of the loss function~\refeq{eq:sf:loss} -- note that all these sub-loss functions contain a regularization strength $C_{\text{?}} > 0$ allowing the user to balance between the different objectives.

\paragraph{Feasibility \& Semifactual property}
As stated in Section~\ref{sec:foundations:semifactuals}, there are two fundamental properties that make up a semifactual explanation: i) the semifactual $\xsf$ must be still rejected but ii) its certainty $\reject(\xsf)$ must not be worse, ideally it should be larger, than the certainty of the original sample $\x$.
We model these two properties as two objectives that are merged together as a weighted sum:
\begin{equation}
\begin{split}
    {\loss}_{\text{feas-sf}}(\xsf)  &= C_\text{feasibility} \cdot \max\left(\reject(\xsf) - \threshold, 0\right) \\
    & + C_\text{sf} \cdot \max\left(\reject(\x) - \reject(\xsf), 0\right)  
\end{split}
\end{equation}

\paragraph{Low complexity}
Since semifactual explanations are feature based explanations, a low-complexity explanation (i.e. easy to understand explanation) should use only very few features. We model this by adding a penalty to the objective if more than $\mu \geq 1$ features are changed:
\begin{equation}\label{eq:sf:simple}
    {\loss}_{\text{simple}}(\xsf)  = C_\text{simple} \cdot \max\left(\sum_{i=1}^{\dimsym} \mathds{1}\big((\xsf - \x)_i \neq 0\big) - \mu, 0\right)
\end{equation}
With the hyperparameter $\mu\geq1$ enables the user to control the complexity of the generated explanations.

\paragraph{Similarity}
While low-complexity is a common requirement of explanations, including semifactual explanations, another special property of semifactuals is that the stated changes must be reasonable large to make the ``even if ...'' explanation useful -- i.e. the semifactual $\xsf$ must be ``sufficiently different'' from the original sample $\x$, which we propose to model by the Euclidean distance:
\begin{equation}\label{eq:sf:similarity}
    {\loss}_{\text{similarity}}(\xsf)  = -C_\text{similarity} \cdot \pnorm{\xsf - \x}_2
\end{equation}
Note the negative sign in front of~\refeq{eq:sf:similarity} which is due to the fact that we minimize the overall loss function~\refeq{eq:sf:loss}. Furthermore, note that~\refeq{eq:sf:similarity} and~~\refeq{eq:sf:simple} are kind of contradictory and for particular scenarios an appropriate balance must be found.

\paragraph{Diversity}
Since the precise meaning of ``diverse semifactuals'' might be different for different use-cases, we propose a very general definition of diversity, namely that the number of simultaneously changed features should be small:
\begin{equation}\label{eq:diversity}
    \diversity({\xsf}^j, {\xsf}^k) = \sum_{i=1}^{\dimsym} \mathds{1}\left(({\deltasf}^j)_i \neq 0 \land ({\deltasf}^k)_i \neq 0\right)
\end{equation}
where ${\deltasf}^j = \xsf^j - \x$ and $\mathds{1}(\cdot)$ denotes the indicator function that returns $1$ if the boolean expression is true and $0$ otherwise. Note that lower values of~\refeq{eq:diversity} correspond to more diverse semifactuals.

Here we propose a sequential approach for computing a set of diverse semifactual explanations: For a set of already computed semifactual explanations $\{\xsf^i\}$ (this set if empty in the beginning), we deduce a set of features $\setF$ that are already used by the current semifactual explanations and therefore should not used again:
\begin{equation}\label{eq:sf:diversity:blacklistedfeatures}
    \setF = \left\{j \mid \exists\,i: (\xsf^i - \x)_j \neq 0\right\}
\end{equation}
Since diverse explanations should not use the same features, we implement a penalty for using already used features:
\begin{equation}\label{eq:sf:diversity}
    {\loss}_{\text{diverse}}(\xsf)  = C_\text{diverse} \cdot \sum_{j\,\in\,\setF}\mathds{1}\Big((\xsf - \x) \neq 0\Big)
\end{equation}
After computing a diverse semifactual explanation, we update~\refeq{eq:sf:diversity:blacklistedfeatures} and compute the next one by solving~\refeq{eq:sf:opt} -- for convenience, we use $\mySF{\x}{\set{F}}{\reject(\cdot)}$ to denote the computation of a semifactual $\xsf$ of a reject option $\reject(\cdot)$ at a given sample $\x$ subject to a set $\set{F}$ of black-listed features.

The complete pseudo-code of computing a set of diverse semifactuals is given in Algorithm~\ref{algo:diverse_sf}.

\begin{table*}
\caption{Quantitative evaluation of our proposed semifactual explanations -- all numbers are rounded to two decimal places.}
\centering
\footnotesize
\begin{tabular}{|c||c||c||c||c||c||}
 \hline
 & \textit{DataSet} & Feas. & Spars. & Div. & Recall \\
 \hline
 \multirow{4}{*}{\rotatebox[origin=c]{90}{k-NN}}
& Wine & $1.0 \pm 0.0$ & $0.08 \pm 0.0$ & $0.0 \pm 0.0$ & $0.14 \pm 0.03$ \\
& Breast cancer & $1.0 \pm 0.0$ & $0.04 \pm 0.0$ & $0.0 \pm 0.0$ & $0.09 \pm 0.0$\\
& Flip & $0.8 \pm 0.16$ & $0.09 \pm 0.0$ & $0.0 \pm 0.0$ & $0.21 \pm 0.03$\\
& t21 & $1.0 \pm 0.0$ & $0.07 \pm 0.0$ & $0.0 \pm 0.0$ & $0.11 \pm 0.01$\\
 \hline\hline
 \multirow{4}{*}{\rotatebox[origin=c]{90}{GNB}}
& Wine & $0.99 \pm 0.0$ & $0.08 \pm 0.0$ & $0.0 \pm 0.0$ & $0.13 \pm 0.04$ \\
& Breast cancer & $0.89 \pm 0.0$ & $0.04 \pm 0.0$ & $0.0 \pm 0.0$ & $0.09 \pm 0.0$\\
& Flip & $0.4 \pm 0.24$ & $0.1 \pm 0.0$ & $0.0 \pm 0.0$ & $0.11 \pm 0.02$\\
& t21 & $0.49 \pm 0.17$ & $0.06 \pm 0.0$ & $0.0 \pm 0.0$ & $0.12 \pm 0.01$\\
 \hline\hline
 \multirow{4}{*}{\rotatebox[origin=c]{90}{RandForest}}
& Wine & $1.0 \pm 0.0$ & $0.08 \pm 0.0$ & $0.0 \pm 0.0$ & $0.11 \pm 0.03$ \\
& Breast cancer & $1.0 \pm 0.0$ & $0.04 \pm 0.0$ & $0.0 \pm 0.0$ & $0.08 \pm 0.0$\\
& Flip & $0.6 \pm 0.24$ & $0.09 \pm 0.0$ & $0.0 \pm 0.0$ & $0.21 \pm 0.04$\\
& t21 & $0.98 \pm 0.0$ & $0.06 \pm 0.0$& $0.02 \pm 0.07$ & $0.11 \pm 0.01$\\
 \hline
\end{tabular}
\label{table:exp:results}
\end{table*}
\section{Experiments}\label{sec:experiments}
We empirically evaluate our proposed mode agnostic semifactual explanations of reject on several ML models and data sets. All experiments are implemented in Python and are publicly available on GitHub\footnote{\url{https://github.com/andreArtelt/DiverseSemifactualsReject}}.

\subsection{Data Sets}
We use the following diverse data sets in our empirical evaluations -- all data sets are standardized.

\subsubsection{Wine}
The ``Wine data set''~\cite{winedata} is used for predicting the cultivator of given wine samples based on their chemical properties. The data set contains $178$ samples and $13$ numerical features such as alcohol, hue and color intensity.

\subsubsection{Breast cancer}
The ``Breast Cancer Wisconsin (Diagnostic) Data Set''~\cite{breastcancer} is used for classifying breast cancer samples into benign and malignant. The data set contains $569$ samples and $30$ numerical features such as area, smoothness and compactness.

\subsubsection{Flip}
This data set is used for the prediction of fibrosis. The set consists of samples of $118$ patients and $12$ numerical features such as blood glucose, BMI and total cholesterol, and was provided by the Department of Gastroenterology, Hepatology and Infectiology of the University Magdeburg~\cite{flipdata}.
As the data set contains some rows with missing values, we chose to replace these missing values with the corresponding feature mean.

\subsubsection{T21}
This data set is used for early diagnosis of chromosomal abnormalities, such as trisomy 21, in pregnant women. The data set consists of $18$ numerical features such as heart rate and weight, and contains over $50000$ samples but only $0.8$ percent abnormal samples (e.g. cases of trisomy 21) -- i.e. it is highly imbalanced. It was collected by the Fetal Medicine Centre at King's College Hospital and University College London Hospital in London~\cite{t21dataset}.

\subsection{Setup}
Since our method Algorithm~\ref{algo:diverse_sf} (see Section~\ref{sec:modeling}) is completely model agnostic, we evaluate it on a set of diverse classifiers: k-nearest neighbors classifier (kNN), Gaussian naive Bayes classifier (GNB), random forest classifier (RandForest). We always use a conformal prediction based implementation of a reject option~\cite{artelt2022model} (see Section~\ref{sec:foundations:rejectoptions}).

For each data sets and each classifier: First, we fit the model and reject option to the data training data. Next, we select a random subset of $30$ percent of the features that are going to be perturbed. We then perturb these features by adding Gaussian noise to the original values. We compute $3$ diverse semifactual explanations (see Section~\ref{sec:modeling}) of those samples that are rejected due to the perturbation -- i.e. samples that were accepted before applying the perturbation, but rejected afterwards. 

We also perform hyperparameter tuning in order to find the best performing model parameters (of the reject option) -- including the hyperparameters of the respective classifiers, which are obtained by a grid search.
In addition, we try to find an appropriate rejection threshold by using the Knee/Elbow method \cite{satopaa2011kneedle}. 
In a real world scenario the threshold might be tuned to allow for a more relaxed or strict rejection scenario, however for the purpose of our research finding the knee point gives us a fairly good approximation of what would usually be considered an appropriate rejection threshold.

Note that we run all experiments in a $5$-fold cross validation.

\subsection{Evaluation}
We evaluate several quantitative aspects of the obtained explanations:
\begin{itemize}
    \item \textit{Feas.}: Feasibility of the explanations -- i.e. whether there are valid semifactual explanations according to Section~\ref{sec:modeling} -- i.e. is the semifactual explanation $\xcf$ still rejected but the assigned certainty $\reject(\xcf)$ is at least as good as the one of the original sample $\x$. Larger numbers are better.
    \item \textit{Spars.}: Sparsity of the explanations -- i.e. how many (percentage) of the available features where changed, smaller numbers are better. Note that sparsity acts as a proxy of ``low-complexity''.
    \item \textit{Div.}: Diversity of the explanations -- i.e. counting the number of simultaneously changed features, smaller numbers are better.
    \item \textit{Recall}: Recall of perturbed features in the explanations -- smaller numbers are ``kind of better'' because semifactual explanations are not supposed to highlight the ``relevant features''.
\end{itemize}
We always report mean and variance of these scores.

\subsection{Results}
The results are shown in Table~\ref{table:exp:results} -- all numbers are rounded to two decimal points.

We observe that our proposed method Algorithm~\ref{algo:diverse_sf}, across different models and data sets, consistently yields sparse and (very) diverse semifactual explanations. Only when applying Gaussian naive Bayes to the flip and t21 data set, our method struggles to compute feasible (i.e. valid) semifactual explanations. Also note that the performance on the flip data set drops for all other methods as well -- we think that the chosen hyperparemeters might not be optimal for this particular data set. Furthermore, we observe that the generated semifactual explanations often do not use the perturbed features -- i.e. these features are the cause/reason of the reject -- this highlights the usefulness of semifactual explanations (i.e. identifying non-relevant features for the ``even if ...'' explanations).

\section{Conclusion}\label{sec:conclusion}
In this work we introduced semifactual explanations for explaining reject options. We proposed the conceptual idea of using diverse semifactual explanations of reject, we proposed a formalization of this concept as well as an implementation of this concept. We empirically evaluated quantitaive aspects of our proposed methods on several data sets and observed a good performance across many different scenarios.

Based on this work, there exist several possible directions for future research:

Our proposed modeling of semifactual explanations of reject is model agnostic -- i.e. it can be applied to any model and reject option. However, exploiting model internals might lead to better performance -- e.g. runtime improvements, formal guarantees, etc. -- and therefore it might be of interest to develop model specific methods for computing semifactual explanations.

The final semifactual explanations as computed by our proposed Algorithm~\ref{algo:diverse_sf} solely depend on the hyperparmeters for balancing between the different objectives as discussed in Section~\ref{sec:modeling}. While it is nice to give the user the ability to ``fine tune'' the explanations, it might be of interest to formalize the semifactuals as a multi-criteria optimization problem and develop a method for automatically exploring (i.e. computing) the pareto front of solutions.

In this work, we ignored the aspect of plausibility -- i.e. we can not guarantee that the semifactual explanations are realistic and plausible in the data domain. However, in particular for example based explanations, missing plausibility might drastically reduce the benefits of the explanation~\cite{kuhl2022keep}. Therefore, future work could address the issue by adding some kind of plausibility constraints to our proposed methods.

Although semifactual explanations are well known in psychology, it is somewhat ``unclear'' how to formalize this concept and consequently compute such explanations. We proposed a reasonable approach that showed good quantitative results. Still it is not clear if our proposed formalization always yields explanations that are useful to humans. Furthermore, it is also unclear if and how these types of explanations are useful for understanding machine learning based decision making systems in general. However, in order to study the usefulness of such explanations, we must first be able to compute such explanations -- which is now possible because of to this work. Consequently, we suggest to evaluate and study the usefulness of our computed semifactuals in a user-study.


\bibliographystyle{IEEEtran}
\bibliography{bibliography}

\end{document}